\documentclass[a4paper, 10 pt, conference]{ieeeconf}  

\IEEEoverridecommandlockouts                              

\usepackage{cite}
\usepackage{amsmath,amssymb,amsfonts}
\usepackage{algorithm}
\usepackage{algpseudocode}
\usepackage[dvips]{graphicx}
\usepackage{textcomp}
\usepackage{xcolor}
\usepackage{bm}
\usepackage{comment}
\usepackage{xcolor}
\usepackage{soul}
\usepackage{url}

\title{\LARGE \bf
Extended Diffeomorphism for Real-Time Motion Replication\\in Workspaces with Different Spatial Arrangements

}

\author{Masaki Saito$^{1,2}$, Shunki Itadera$^{1}$ and Toshiyuki Murakami$^{2}$
\thanks{This work was supported in part by JSPS KAKENHI JP24K20826 and JST CRONOS JPMJCS24K6.}
\thanks{$^{1}$National Institute of Advanced Industrial Science and Technology (AIST),
        Tokyo, Japan
        {\tt\small (e-mail: \{saitou-aist312, s.itadera\}@aist.go.jp)}}%
\thanks{$^{2}$Department of System Design Engineering, Keio University,
        Yokohama, Japan
        {\tt\small (e-mail: mura@sd.keio.ac.jp)}}%
}

\begin{document}

\maketitle
\thispagestyle{empty}
\pagestyle{empty}

\begin{abstract}
This paper presents two types of extended diffeomorphism designs to compensate for spatial placement differences between robot workspaces. Teleoperation of multiple robots is attracting attention to expand the utilization of the robot embodiment.
Real-time reproduction of robot motion would facilitate the efficient execution of similar tasks by multiple robots. A challenge in the motion reproduction is compensating for the spatial arrangement errors of target keypoints in robot workspaces. This paper proposes a methodology for smooth mappings that transform primary robot poses into follower robot poses based on the predefined key points in each workspace. Through a picking task experiment using a dual-arm UR5 robot, this study demonstrates that the proposed mapping generation method can balance lower mapping errors for precise operation and lower mapping gradients for smooth replicated movement.
\end{abstract}

\section{INTRODUCTION}

\label{introduction}
Robotic teleoperation plays an important role in hazardous environments and in areas that are inaccessible to humans.
A fundamental functionality of a teleoperation system is to map the operator's input to the robot's movement.
When a single operator controls multiple robots to perform a collaborative task, such as lifting a large object, the operator needs to repeat similar input to control each robot in the non-common action phase, e.g., guiding individual robot to the corresponding grasping point.
Once the geometric errors among the robots become acceptable, the operator can send a common command to control all the robots simultaneously, e.g., lifting the table up.

Although multi-robot teleoperation is a promising method to expand the usage of robot embodiment, large workloads are required for precise control in the non-common action phase.
Therefore, a reduction method of operational workload is important for practical multi-robot teleoperation. 
One approach is to develop an interface to easily select a target robot in the non-common action phase and a shared control surface in the common action phase\cite{Odzmar}. 
While this approach reduces the operation workload for selecting robots,
the operator still needs to control robots individually, which would lead to a decrease in work efficiency.
In this study, we address a methodology to control multiple robots simultaneously to improve the efficiency in the non-common action phase.
This paper tackles on real-time replication of a primary robot action for the other robots, considering the geometric difference among the robots' workspaces.
This approach aims to allow the operator to focus on and control a primary robot, and the other robots automatically adjust the geometric errors and reach the target poses without time delay.

\begin{figure}[t]
\begin{center}
  \includegraphics[width=0.8\linewidth]{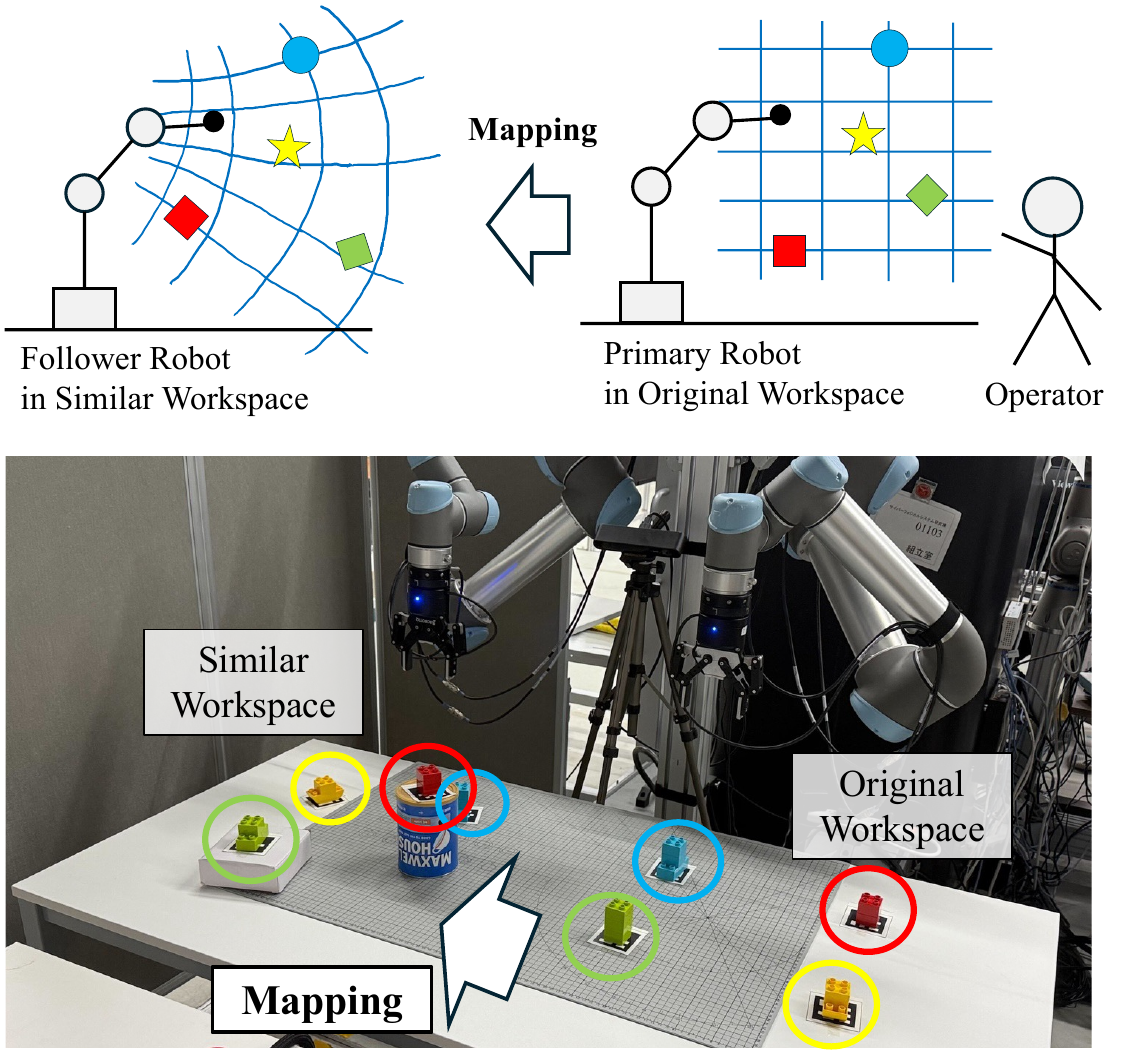}
  \caption{Concept of our proposed diffeomorphism mapping.}
  \label{fig:concept_images} 
\end{center}
\end{figure}

A notable example of non-real-time motion replication is imitation learning (IL).
IL is an effective method for generating the movements of a robot in a replicated environment from the movements of a reference robot. 
For instance, Dynamic Movement Primitives (DMP) can extract the essence of the robot's motion from a single human demonstration and adjust it in another workspace\cite{DMP1}.
In situations where a user operates multiple robots, IL is not suitable because the goal pose is not given in advance.

There have been several studies on real-time motion replication for simultaneous multi-robot operation.
For instance, 
variable scaling methods \cite{scaling_kobayashi,scaling_rate} are used to adjust motion scaling factors for each robot to absorb geometric error in the corresponding environment. 
However, such scaling methods require updating the scaling coefficients when the geometric configuration of the workspace changes, leading to a discontinuous and non-transparent experience. 
To overcome the shortcomings of the scaling methods, a diffeomorphism-based spatial mapping method has been proposed \cite{Gao}.
The map is represented by a weighted composition function of radial basis functions (RBF), which is continuous and differentiable, and designed for minimizing the inter-environmental errors of position and orientation between the robots.
However, in highly unstructured environments, the author reported that significant distortion occurs in the mapping due to limited compensation for large inter-environmental errors. Since excessive distortion in positional mapping causes a large robot velocity, the gradient of mapping needs to be low in practice. Nevertheless, the conventional methods are hard to achieve both a lower mapping gradient and a lower mapping error in some cases.

This paper aims to define a novel diffeomorphism to minimize both mapping gradient and mapping error, even in cases of highly geometric randomized workspaces.
We propose two types of extended diffeomorphisms: Rotation Extended Diffeomorphism and Twisted Affine Extended Diffeomorphism. Rotation Extended Diffeomorphism is an extension of vector-based positional diffeomorphism to quaternion-based one. Twisted Affine Extended Diffeomorphism is an extension of the conventional method with the addition of a non-linear mapping for workspaces with various deformations.
Fig. \ref{fig:concept_images} shows a concept image of our proposed mapping methodology. We demonstrate the effectiveness of the presented methodology through simulation and experiments.

\section{ Diffeomorphism for Motion Replication}
\label{mapping}

We define a mapping formulation for real-time motion replication.
The transformation from the primary robot position $\bm{p}$ and orientation $\bm{q}$ to another robot's position $\hat{\bm{p}}$ and orientation $\hat{\bm{q}}$ is represented as follows:
\begin{align}
    \hat{\bm{p}} = \hat{\bm{\Phi}} (\bm{p}), \ \hat{\bm{q}} = \hat{\bm{\Psi}}(\bm{p}, \bm{q}),
    \label{eq:mapping_func}
\end{align}
where $\hat{\bm{\Phi}}$ and $\hat{\bm{\Psi}}$ are diffeomorphic mapping functions for position and orientation, respectively.
This paper address a novel design of the position mapping function ${\bm{\hat{\Phi}}}$.

\subsection{Conventional Diffeomorphism}
According to the prior work \cite{Gao}, we here define conventional diffeomorphic mapping functions $\hat{\bm{\Phi}}$ and $\hat{\bm{\Psi}}$ in eq.\,(\ref{eq:mapping_func}) as $\hat{\bm{\Phi}}_{\rm DIFF}$ and $\hat{\bm{\Psi}}_{\rm DIFF}$, which are written as:
\begin{align}
    \hat{\bm{\Phi}}_{\rm DIFF}(\bm{p}) &= \bm{\Phi}_{J}\circ\bm{\Phi}_{J-1}\circ\cdots\circ\bm{\Phi}_2\circ\bm{\Phi}_1(\bm{p}), \label{total_position} \\
    \hat{\bm{\Psi}}_{\rm DIFF}(\bm{p}, \bm{q}) &= \bm{\Psi}_{J}\circ\bm{\Psi}_{J-1}\circ\cdots\circ\bm{\Psi}_2\circ\bm{\Psi}_1(\bm{p}, \bm{q}), \label{total_posture} 
\end{align}
where $\bm{\Phi}_{j}$ and $\bm{\Psi}_{j}$ ($j=1\cdots J$) are $j$-th basic composition function for position and orientation mapping, and $J$ is the total number of the composition.
Note that the forms of the basic composition functions are unique, and we omit the $j$ index for a simple expression.
The basic composition functions ($\bm{\Phi}(\bm{p})$ and $\bm{\Psi}(\bm{p}, \bm{q})$) are defined as:
\begin{align}
\bm{\Phi}(\bm{p}) &=\bm{p}+k_f(\bm{p}| \rho_{1}, \bm{c}_{1})\cdot \bm{v}_{1}, \label{j-th_position} \\
  \bm{\Psi}(\bm{p}, \bm{q}) &={\bm{v}_{2}}^{k_f(\bm{p} | \rho_{2}, \bm{c}_{2})} \otimes \bm{q}, \label{j-th_posture}
\end{align}
where $\bm{v}_1 \in \mathbb{R}^3$ and $\bm{v}_2 \in \mathbb{H}$ are corresponding to the amount of linear transformation of key points in position and orientation.
The transformation is weighted by RBF, which is represented by
\begin{align}
k_f(\bm{p} | \rho, \bm{c}) = \exp(-\rho^2||\bm{p}-\bm{c}||^2), \label{equ:rbf}
\end{align}
where $\rho \in \mathbb{R}^1$ and $\bm{c} \in \mathbb{R}^3$ are variables determining the shape of the function, corresponding to the slope and center position of RBF, respectively.
The previous study in \cite{Gao} presented a sufficient condition for diffeomorphism as:
\begin{align}
{0<\rho_1<\frac{1}{\sqrt{2}||\bm{v}_1||}e^{\frac{1}{2}}}.
\end{align}

While conventional mapping can absorb translational error, it is not compatible with workspace rotation and reflection cases. That is because the optimization parameter ${\bm{v}_1}$ is used for a locally weighted shift.
In this paper, we present two extended diffeomorphisms to address the rotation, reflection, and swapping of key points in workspaces.

\subsection{Rotation Extended Diffeomorphism}
\subsubsection{Orbital-Rotation for Error Representation}
\begin{figure}[t]
\begin{center}
  \includegraphics[width=0.8\linewidth]{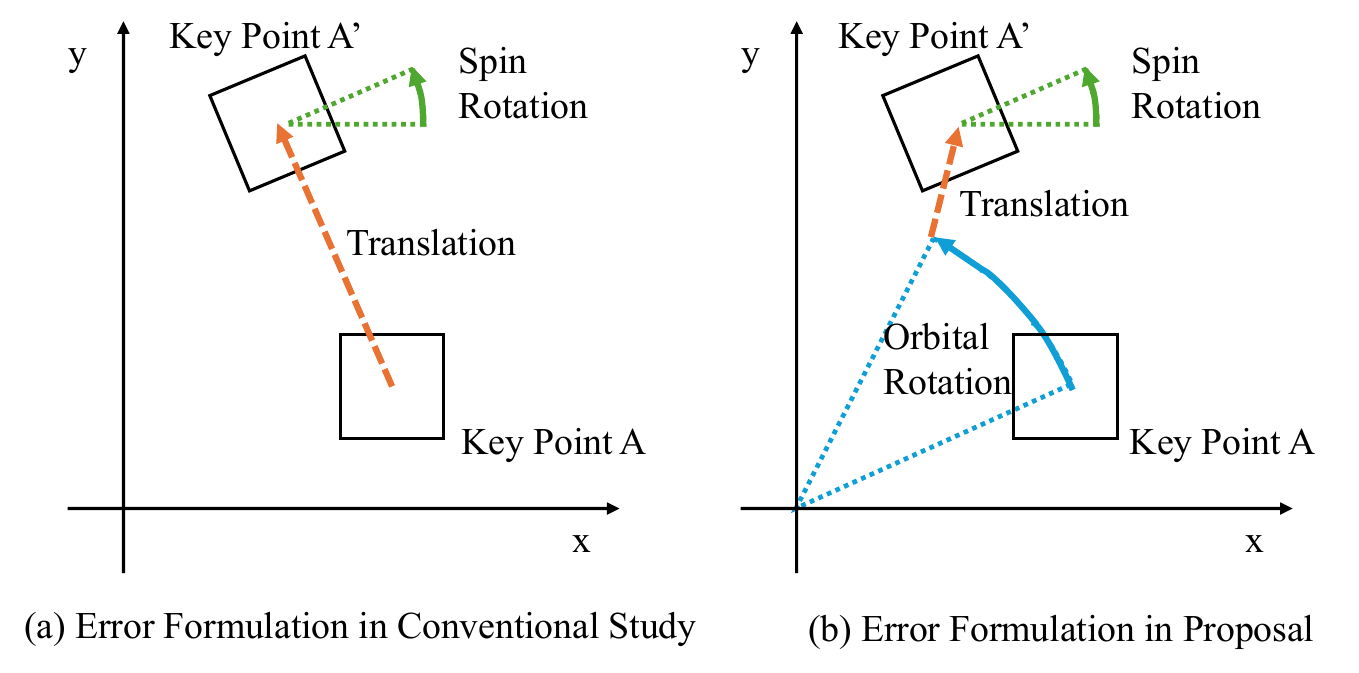}
  \caption{Geometric error formulation among workspaces.}
  \label{fig:error_formulation} 
\end{center}
\end{figure}
\begin{figure}[t]
\begin{center}
  \includegraphics[width=1.0\linewidth]{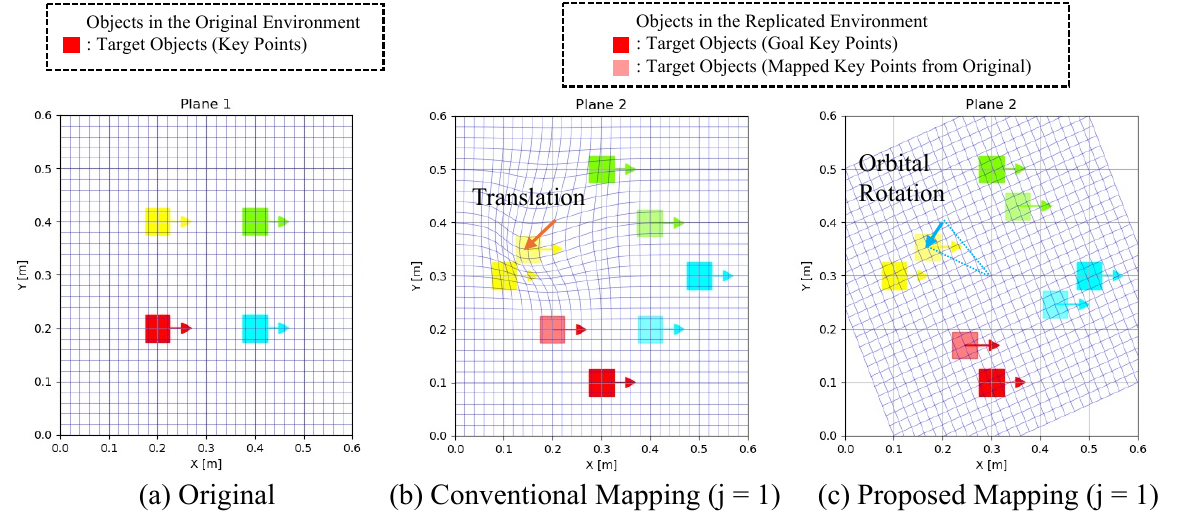}
  \caption{Comparison of the initial mapping between DIFF and R-DIFF.}
  \label{fig:tutorial_mapping} 
\end{center}
\end{figure}
In order to improve the compatibility of the conventional diffeomorphic mapping with rotated situations, we redefine the geometric error formulation among the workspaces.
Fig. \ref{fig:error_formulation} shows the error formulation when key point A shifts and rotates as key point A'. 
The conventional method shown in Fig. \ref{fig:error_formulation}(a) represents the geometric error as translation and spin-rotation terms.
Since consideration of the rotation of the key points group is not sufficient, we additionally consider the orbital-rotation term for position error, which provides a novel form of diffeomorphism compatible with orbital-rotational error as shown in Fig. \ref{fig:tutorial_mapping}.
In the mathematical expression, we define the diffeomorphic function $\hat{\bm{\Phi}}$ in eq.\,(\ref{eq:mapping_func}) as $\hat{\bm{\Phi}}_{\rm R-DIFF}$ and modify the basic composition function for position in eq.\,(\ref{j-th_position}) as follows:
\begin{align}
\bm{\Phi}(\bm{p}) &= {\bm{v}_{3}}^{k_f(\bm{p}|\rho_3, \bm{c}_{3})}
  \cdot \bm{p^q}
  \cdot \bar{\bm{v}}_{3}^{k_f(\bm{p}|\rho_3, \bm{c}_{3})}+k_f(\bm{p}| \rho_{1}, \bm{c}_{1})\cdot \bm{v}_{1},\label{j-th_position_q}
\end{align}
where $\bm{p^q}$ is a pure quaternion form of ${\bm{p}}$.
${\bm{v}_{3}}$ is a quaternion representing the orbital-rotational error of key points.

\subsubsection{Parameter Optimization}
We present an algorithm for determining the parameters of our diffeomorphic mapping as shown in Algorithm \ref{alg:diffeomorphic}.
The sets of variable parameters for position and orientation mapping are composed of ${ \rho_{1,j}, \bm{c}_{1,j}, \bm{v}_{1,j}, \rho_{3,j}, \bm{c}_{3,j}, \bm{v}_{3,j} }$ ($j=1\cdots J$) and ${\rho_{2,j}, \bm{c}_{2,j}, \bm{v}_{2,j} }$ ($j=1\cdots J$), respectively.
In the optimization process, parameters ${\bm{c}, \bm{v}}$ are selected based on a geometric relationship. After deciding ${\bm{c}, \bm{v}}$, a parameter ${\rho}$ is selected to minimize the following cost functions:
\begin{align}
F_{\Phi} (\rho_{1,j}, \rho_{3,j}, \bm{v}_{1,j}, \bm{v}_{3,j})
  & = \frac{1}{K} \sum_{k = 1}^K 
  \Big\| \bm{p}_{k, {\rm goal}} - \Phi_j(\bm{p}_{k,j})\Big\|, \label{F_position}\\
F_{\Psi} (\rho_{2,j}, \bm{v}_{2,j})
  &=\frac{1}{K} \sum_{k = 1}^K 
  \Big\| \log(\bm{q}_{k, {\rm goal}} \otimes \overline{\Psi_j(\bm{q}_{k,j}) }) \Big\|, \label{F_orientation}
\end{align}
where $\bm{p}_{k, {\rm goal}}$ denotes the goal position of key points, $\bm{q}_{k, {\rm goal}}$ denotes the goal orientation of key points, and $\bm{r}_{k, {\rm goal}}$ denotes the goal orbital-rotation of key points. $k$ is the index of key points.
 ${\bm{q}_0}$ is a null rotation quaternion.
The composition iteration of RBF is terminated when the position and orientation errors fall below a predefined minimum threshold, or when the number of compositions $j$ reaches the maximum $J_{max}$.

\begin{algorithm}[H]
\caption{Diffeomorphism Parameter Optimization}
\label{alg:diffeomorphic}
\begin{algorithmic}[1]
\Require 
$\bm{P}_\mathrm{inits}=\{\bm{p}_{k, \mathrm{init}}, \bm{q}_{k, \mathrm{init}}, \bm{r}_{k, \mathrm{init}}\}_{k=1}^K$, 
$\bm{P}_\mathrm{goals}=\{\bm{p}_{k, \mathrm{goal}}, \bm{q}_{k, \mathrm{goal}}, \bm{r}_{k, \mathrm{goal}}\}_{k=1}^K$
\Ensure 
$\bm{\rho}=\{\bm{\rho}_{1,j}, \bm{\rho}_{2,j}, \bm{\rho}_{3,j}\}_{j=1}^J$, 
$\bm{C}=\{\bm{c}_{1,j}, \bm{c}_{2,j}, \bm{c}_{3,j}\}_{j=1}^J$, 
$\bm{V}=\{\bm{v}_{1,j}, \bm{v}_{2,j}, \bm{v}_{3,j}\}_{j=1}^J$
\State Initialize $\bm{P}_j \gets \bm{P}_\mathrm{inits}$
\For{error (\ref{F_position}), (\ref{F_orientation}) small enough}
    \State ${m := \arg \max(||\bm{p}_{k, j} - \bm{p}_{k, \text{goals}}||)},\  _{k \in [1, \cdots, K]}$ 
    \State ${n := \arg \max(||\log(\bm{q}_{k, j} * \overline{\bm{q}_{k, \text{goals}}})||)},\  _{k \in [1, \cdots, K]}$
    \State ${o := \arg \max(||\log(\bm{r}_{k, j} * \overline{\bm{r}_{k, \text{goals}}})||)},\  _{k \in [1, \cdots, K]}$
    \State $ \bm{c}_{1,j} , \bm{c}_{2,j}, \bm{c}_{3,j} \gets \bm{p}_{m, j}, \bm{p}_{n, j}, \bm{p}_{o, j} $
    \State $ \bm{v}_{1,j} , \bm{v}_{2,j}, \bm{v}_{3,j} \gets (\bm{p}_{m, \mathrm{goal}}-\bm{p}_{m,j}), (\bm{q}_{n, \mathrm{goal}} * \overline{\bm{q}_{n, j}}), (\bm{r}_{o, \mathrm{goal}} * \overline{\bm{r}_{o, j}}) $
    \State Optimize $\bm{\rho}_{1,j}$ to minimize  ${F_{{\Phi}}(\rho_{1,j}, 0, \bm{v}_{1,j}, \bm{q}_0)}$
    \State Optimize $\bm{\rho}_{2,j}$ to minimize  ${F_{{\Psi}}(\rho_{2,j}, \bm{v}_{2,j})}$
    \State Optimize $\bm{\rho}_{3,j}$ to minimize  ${F_{{\Phi}}(0, \rho_{3,j}, \bm{0}, \bm{v}_{3,j})}$
    \If {${F_\Phi(0, \rho_{3,j}, \bm{0}, \bm{v}_{3,j}) < F_{{\Phi}}(\rho_{1,j}, 0, \bm{v}_{1,j}, \bm{q}_0)}$}
        \State $ \rho_{1,j} , \bm{v}_{1,j} \gets 0 , \bm{0} $
    \Else
        \State $ \rho_{3,j} , \bm{v}_{3,j} \gets 0 , \bm{q}_0 $
    \EndIf
    \State $\bm{p}_{k, j+1}, \bm{q}_{k, j+1} = \Phi_j(\bm{p}_{k,j}), \Psi_j(\bm{q}_{k,j})$
\EndFor
\end{algorithmic}
\end{algorithm}

\subsection{Twisted Affine Extended Diffeomorphism}
\subsubsection{Twisted Affine Transformation}
\begin{figure}[t]
\begin{center}
  \includegraphics[width=1.0\linewidth]{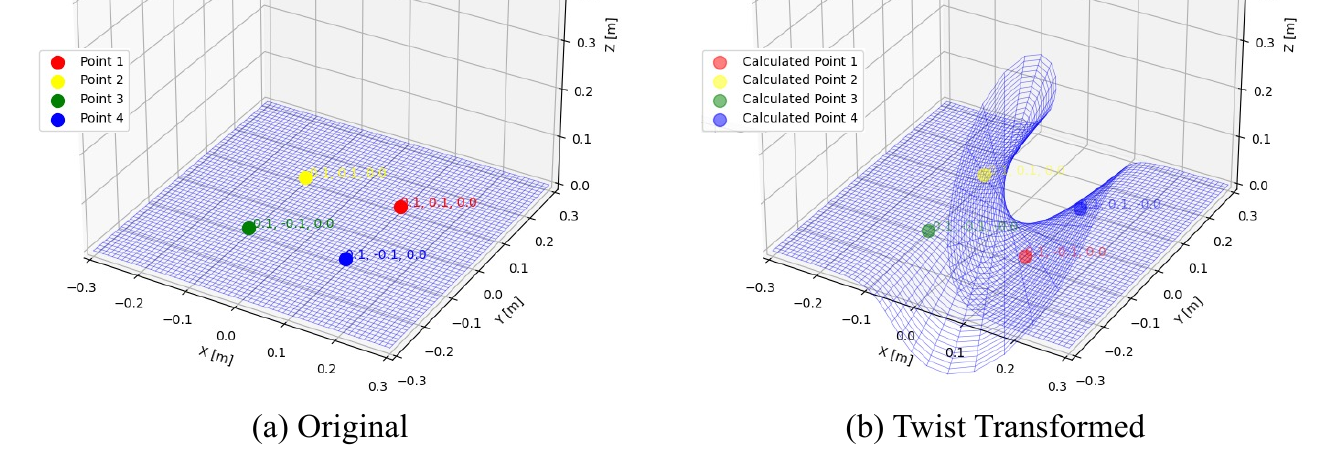}
  \caption{Our presented twist transformation of the xy-plane.}
  \label{fig:twist} 
\end{center}
\end{figure}

In this paper, assuming that there are four key points, we present a twist transformation to address key points swapping. 
For instance, the twist transformation of the roll axis can be defined using a rotation matrix in the roll direction ${R_{\text{roll}}}$ as:
\begin{align}
\bm{\hat{p}}
 &= 
R_{\text{roll}}(\phi_t) \cdot
\bm{p} , \label{eq:twist} 
\end{align}
where the twist angle ${\phi_t(x)}$ is a sigmoid function, which can be written as:
\begin{align}
\phi_t(x) 
&= \phi_{t0} \cdot \frac{1}{1 + e^{-\rho_{\mathrm{sig}} x}}, \label{equ:sigmoid} 
\end{align}
where ${\rho_{\mathrm{sig}}}$ is a parameter of the intensity of the twist.
The result of applying the twist transformation of the roll axis to the xy-plane is shown in Fig. \ref{fig:twist}. 
In this paper, the twist transformation is defined in roll and pitch directions.

By combining the twist transformation with the affine transformation, the mapping can address key points' scaling, rotation, reflection, and swapping simultaneously.
We present the mapping method as Twisted Affine Extended Diffeomorphism, written as:
\begin{align}
\hat{\bm{\Phi}}_{\rm TA}(\bm{p})
&= R_{\rm TA} \bm{p} + \bm{t}_{\rm TA} ,\label{eq:twisted_affine} 
\end{align}
where
\begin{align}
R_{\text{TA}} &= R_{\text{scaling}}(a_x, a_y, a_z) \cdot R_{\text{shear}}(s_{xy}, s_{xz}, s_{yz}) \notag\\
 &\cdot R_{\text{rotation}}(\phi, \theta, \psi) \cdot R_{\text{reflection}}(r_x, r_y, r_z) \notag \\
 &\cdot R_{\text{twist}}(\phi_t(x), \theta_t(y)) ,\label{eq:twisted_affine_rotation} \\
\bm{t}_{\text{TA}} &= 
\left[
\begin{array}{ccc}
t_x& t_y& t_z
\end{array}
\right]^T . \label{equ:twisted_affine_t} 
\end{align}
The definitions of each rotation matrix are provided in the Appendix. The parameters of the Twisted Affine mapping are shown in Table \ref{params_twisted_affine}. 

\begin{table}[htbp]
\caption{Parameters of the Twisted Affine Transformation}
\begin{center}
\begin{tabular}{ccc}
\hline\hline
Parameter & Meaning & Range \\
\hline
$a_x, a_y, a_z $ & Scaling along $x$, $y$, $z$ axes & $ [0.5, 2] $ \\
$s_{xy}, s_{xz}, s_{yz}$ & Shear in $xy$, $xz$, $yz$ planes & $ [0, 1]$ \\
$\phi, \theta, \psi$ & Rotation around $x$, $y$, $z$ axes & $ [-\frac{\pi}{2}, \frac{\pi}{2}]$ \\
$r_x, r_y, r_z$ & Reflection along $x$, $y$, $z$ axes & $ \{-1, 1\}$ \\
$\phi_t, \theta_t$ & Twist angles around $x$, $y$ axes & $ \{0, \pi\}$ \\
$t_x, t_y, t_z$ & Translation vector & $ -- $ \\
\hline\hline
\end{tabular}
\label{params_twisted_affine}
\end{center}
\end{table}

\subsubsection{Parameter Optimization}
Algorithm \ref{alg:twisted_affine} shows the optimization process. The discrete parameters ${\bm{B}}$ are selected in order. The continuous parameters ${\bm{A}}$ are selected with the BFGS method to minimize the following cost functions:
\begin{align}
  F_{\rm TA} 
  & = \frac{1}{K} \sum_{k = 1}^K 
  \Big\| \bm{p}_{k, {\rm goal}} - \Phi_{\rm TA}(\bm{p}_{k, {\rm init}})\Big\| . \label{F_twisted_affine}
\end{align}
\begin{algorithm}[H]
\caption{Twisted Affine Parameter Optimization}
\label{alg:twisted_affine}
\begin{algorithmic}[1]
\Require 
$\bm{P}_\mathrm{inits}=\{\bm{p}_{k, \mathrm{init}}, \bm{q}_{k, \mathrm{init}}\}_{k=1}^K$, 
$\bm{P}_\mathrm{goals}=\{\bm{p}_{k, \mathrm{goal}}, \bm{q}_{k, \mathrm{goal}}\}_{k=1}^K$
\Ensure 
$\bm{A}=\{a_x, a_y, a_z, s_{xy}, s_{xz}, s_{yz}, \phi, \theta, \psi\}$
$\bm{B}=\{r_x, r_y, r_z, \phi_{t0}, \theta_{t0}\}$
$\bm{T}=\{t_x, t_y, t_z\}$
\State $\bm{T} \gets \sum_{k=1}^K (\bm{p}_{k, \mathrm{goal}})  - \sum_{k=1}^K(\bm{p}_{k, \mathrm{init}})$
\For{$\bm{B}$ choose candidates from TABLE \ref{params_twisted_affine}}
    \State Optimize ${\bm{A}}$ to minimize positional error ${F_{\rm TA}}$
    \If {${F_{\rm TA}} < best\_objective$}
        \State $best\_objective \gets {F_{\rm TA}}$
        \State $ \bm{A}\_opt ,\bm{B}\_opt  \gets {\bm{A}}, {\bm{B}}$
    \EndIf
\EndFor
\State $\bm{A}, \bm{B} \gets \bm{A}\_opt, \bm{B}\_opt$
\end{algorithmic}
\end{algorithm}

\section{Manipulator Control}
\label{controller_design}
We simultaneously control multiple robots, which are placed in geometrically randomized workspaces, using our presented motion replication method.
We input a target end-effector pose of the primary robot using a teleoperation interface.
The other robots follow transformed target poses calculated from eq.\,(\ref{eq:mapping_func}).
Then we obtain an augmented variable $\bm{x}_i$ of the target pose of $i$-th robot as $\bm{x}_i = [\bm{p}_i^T\ \bm{q}_i^T]^T$

The robot controller to follow the target poses is implemented using OpenHRC \cite{OpenHRC}.
We apply a proportional control shown in eq.\,(\ref{eq:velocity_control}) and use its output as target end-effector velocities.
\begin{align}
    \bm{\dot{x}}_i = k_p(\bm{x}_i - \bar{\bm{x}}_i), \label{eq:velocity_control}
\end{align}
where $\bar{\bm{x}}_i$ denotes the current pose of the $i$-th robot.
${k_p = 2.0}$ is the proportional gain.
The inverse kinematics that converts the desired end-effector velocity to the desired joint velocity is formulated as a QP (quadratic programming) optimization problem as shown in eq.\,(\ref{equ:IK}).
\begin{align}
  \min_{\bm{\dot{\theta}}_i} 
  \sum_{i}^{N} 
  &\left( \bigl| \bm{J}_i(\bm{\theta}_i){\bm{\dot{\theta}}_i} - \bm{\dot{x}}_i \bigr|^{2} +|\mathbf{\bm{\dot{\theta}}_i}|^2 \right),
  \label{equ:IK}
\end{align}
where $N$ is the number of robots.
$\bm{\theta}_i$ and ${\bm{J}_i(\bm{\theta})}$ is the joint angle and Jacobian matrix of $i$-th robot, respectively.
The obtained joint velocity is sent to the robot via a Joint Velocity Controller.

\section{Simulation}
\begin{figure}[t]
\begin{center}
  \includegraphics[width=1.0\linewidth]{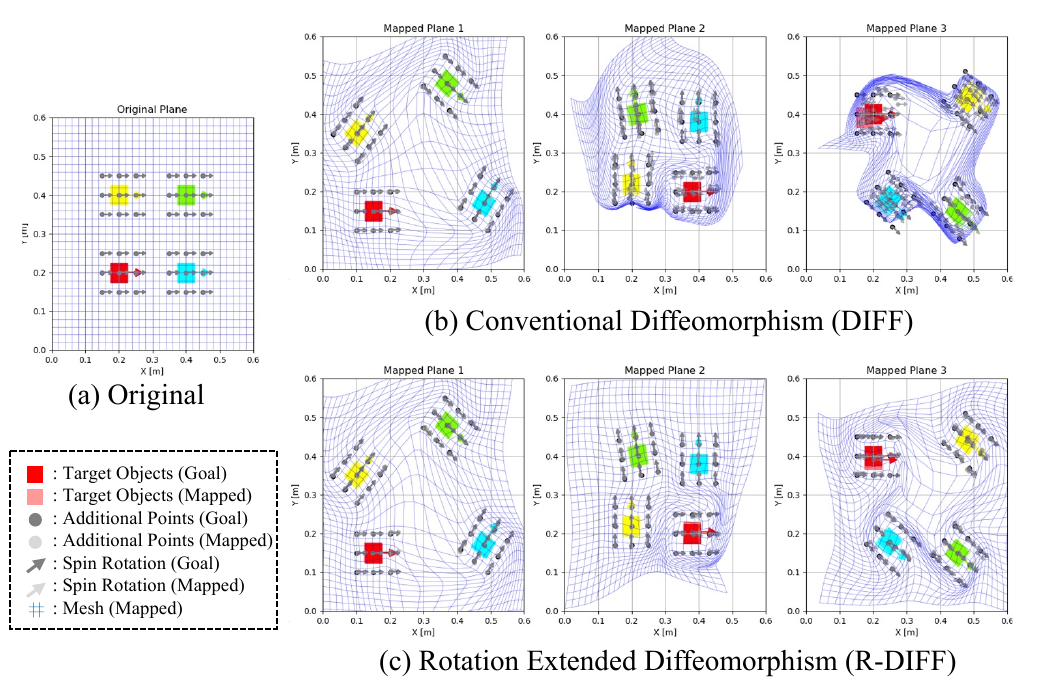}
  \caption{Simulation1: Comparison of the mapping shape between DIFF and R-DIFF.}
  \label{fig:sim1_result}
\end{center}
\end{figure}
\subsubsection{Simulation 1}
We conduct numerical simulations to compare the conventional Diffeomorphism (DIFF) and the proposed Rotation Extended Diffeomorphism (R-DIFF). In this simulation, multiple sub-key points around target objects are added to improve the consistency of geometric information around targets. The number of target objects $K$ is set to 4.
Sub key points around each target object are placed from a distance of 0.05\,m in x,y,z directions.
The total number of sub-key points ${K_s}$ is 27. All points, including target objects and sub-key points around target objects, are used as key points ${\bm{P}_\mathrm{inits}, \bm{P}_\mathrm{goal}}$ in Algorithm 1.  The number of replicated environments is 3. The maximum number of RBF iterations $J_{max}$ is set to 100. Iteration ends when eq.\,(\ref{F_position}) is under 5\,mm.

Fig. \ref{fig:sim1_result} shows the results of simulation 1. In the case of DIFF, a steep gradient and positional error are confirmed in Planes 2 and 3. However, in the case of R-DIFF, the gradient is gentle and the positional error is nearly zero. In Plane 3, the positional error of the key points is 18.65\,mm (DIFF), 5.58\,mm (R-DIFF). 

\begin{figure}[t]
\begin{center}
  \includegraphics[width=1.0\linewidth]{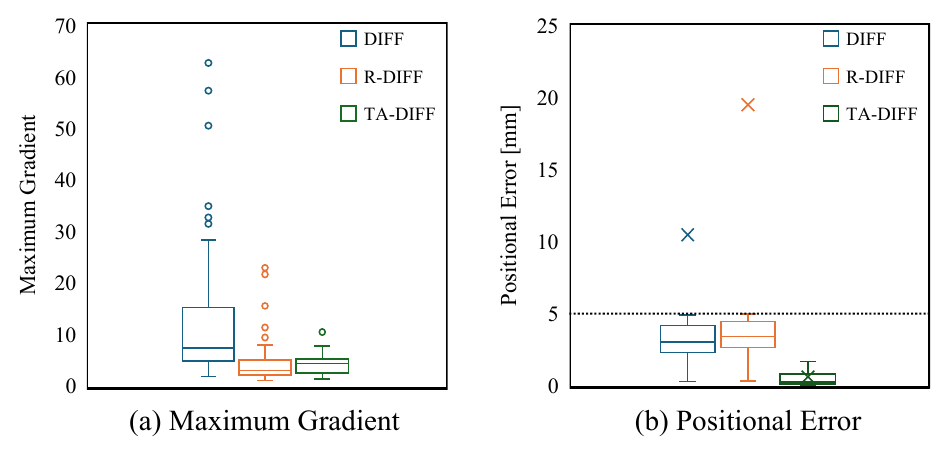}
  \caption{Simulation2: Comparison of evaluation index between DIFF, R-DIFF, and TA-DIFF.}
  \label{fig:sim_gradient}
\end{center}
\end{figure}

\begin{table}[t]
\caption{Evaluation metrics in Simulation 2}
\begin{center}
\begin{tabular}{cccc}
\hline\hline
 & (a) DIFF& (b) R-DIFF& (c) TA-DIFF\\
\hline
Positional Error (Max)& 145\,mm & 211\,mm & 4.73\,mm \\
Success Rate & 90\,\% & 87\,\% & 100\,\% \\
RBF Iteration (Average)& 39 & 29 & 10 \\
\hline\hline
\end{tabular}
\label{tab:sim_mapping}
\end{center}
\end{table}

\subsubsection{Simulation 2}
Simulation 2 examines the reliability of the mapping in various situations. DIFF, R-DIFF, and Twisted Affine Diffeomorphism (TA-DIFF) are compared in the situation where target objects are placed randomly within the range ${\bm{p}_{k, {\rm goal}}\in[0.1, 0.6]^3}$. There are no sub-key points around target objects ${(K_s=0)}$. The number of replicated environments $N$ is 100. The other conditions are the same as ones in simulation 1.\par
Fig. \ref{fig:sim_gradient} is the boxplot showing the maximum gradient and positional error between target objects. The maximum gradient is calculated by the spectral norm of the Jacobian matrix for each grid point set at 0.02\,m intervals on the map. The positional error between target objects is calculated using the value of eq.\,(\ref{F_position}) at the end of the mapping algorithm.
From the results, the proposed method was confirmed to generate smoother mappings with smaller velocity gradients, taking into account the geometric relationships between target objects. Regarding positional errors, TA-DIFF was confirmed to be less than 5\,mm in all situations, but R-DIFF showed larger positional errors. This is thought to be because eq. \,(\ref{j-th_position_q}) does not provide appropriate compensation in some cases.
Table \ref{tab:sim_mapping} summarizes the other evaluation values for each method. In DIFF and R-DIFF, there were approximately 10\,\% cases where the error did not fall below the desired value after 100 RBF superpositions. However, the proposed TA-DIFF guaranteed that the position error in the mapped environment would always fall below the target minimum error under all conditions.

\begin{figure}[t]
\centering
  \includegraphics[width=1.0\linewidth]{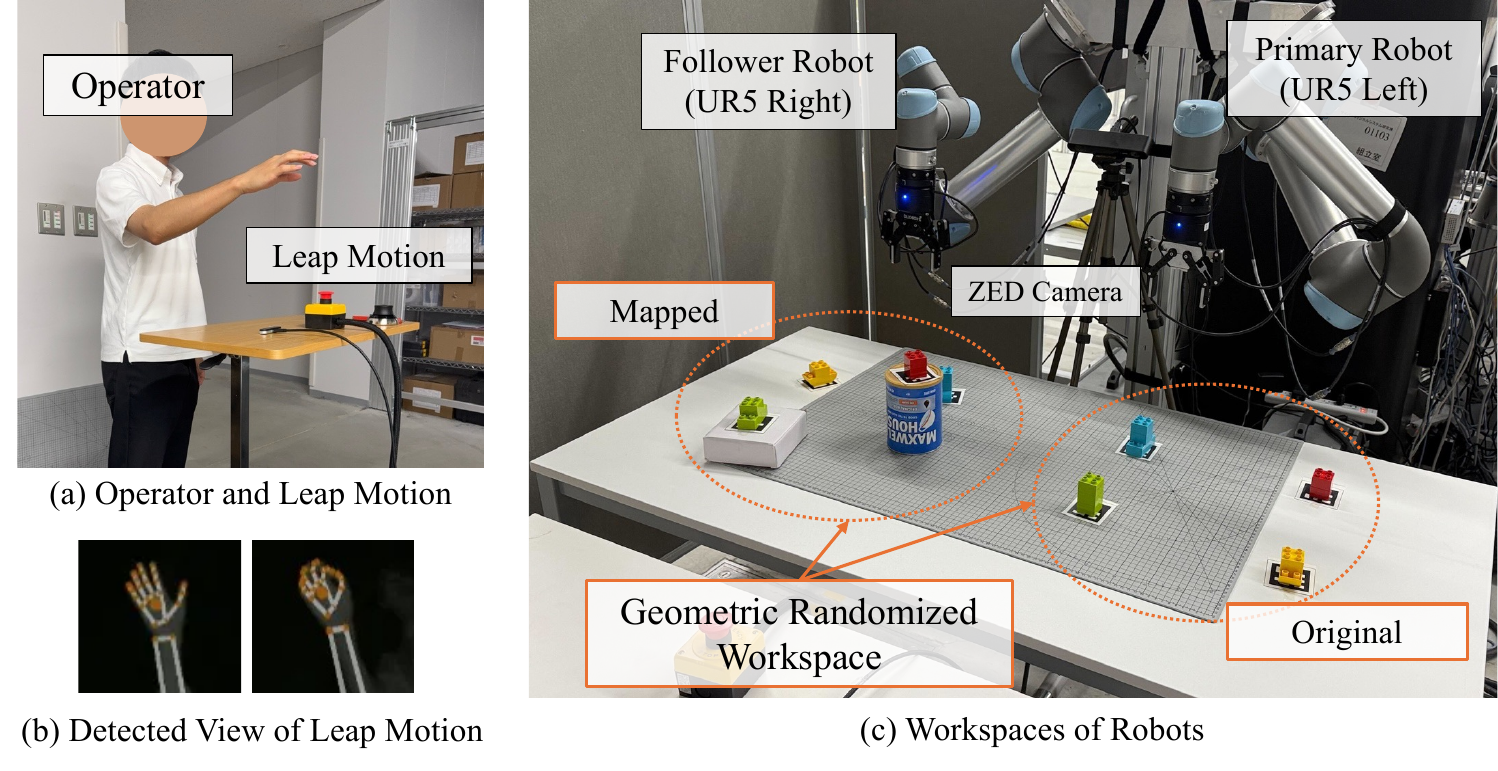}
  \caption{Experimental setup.}
  \label{fig:experimental_setup}
\end{figure}
\begin{figure}[t]
  \includegraphics[width=1.0\linewidth]{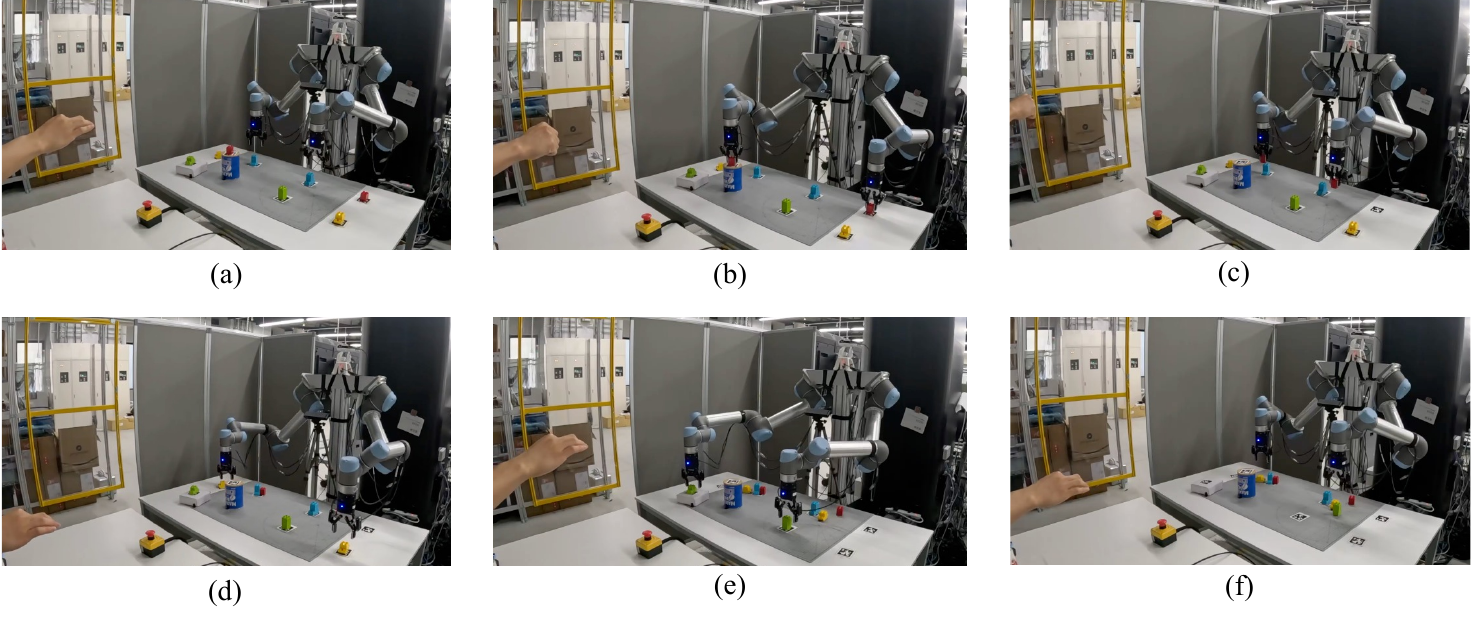}
  \caption{Task flow in Experiment 2. (a) Initial state (b) Picking red target (c) Carrying red target (d) Picking yellow target (e) Picking green target (f) Finished all picking task.}
  \label{fig:exp2-result}
\end{figure}
\begin{figure*}[t]
\begin{center}
  \includegraphics[width=1.0\linewidth]{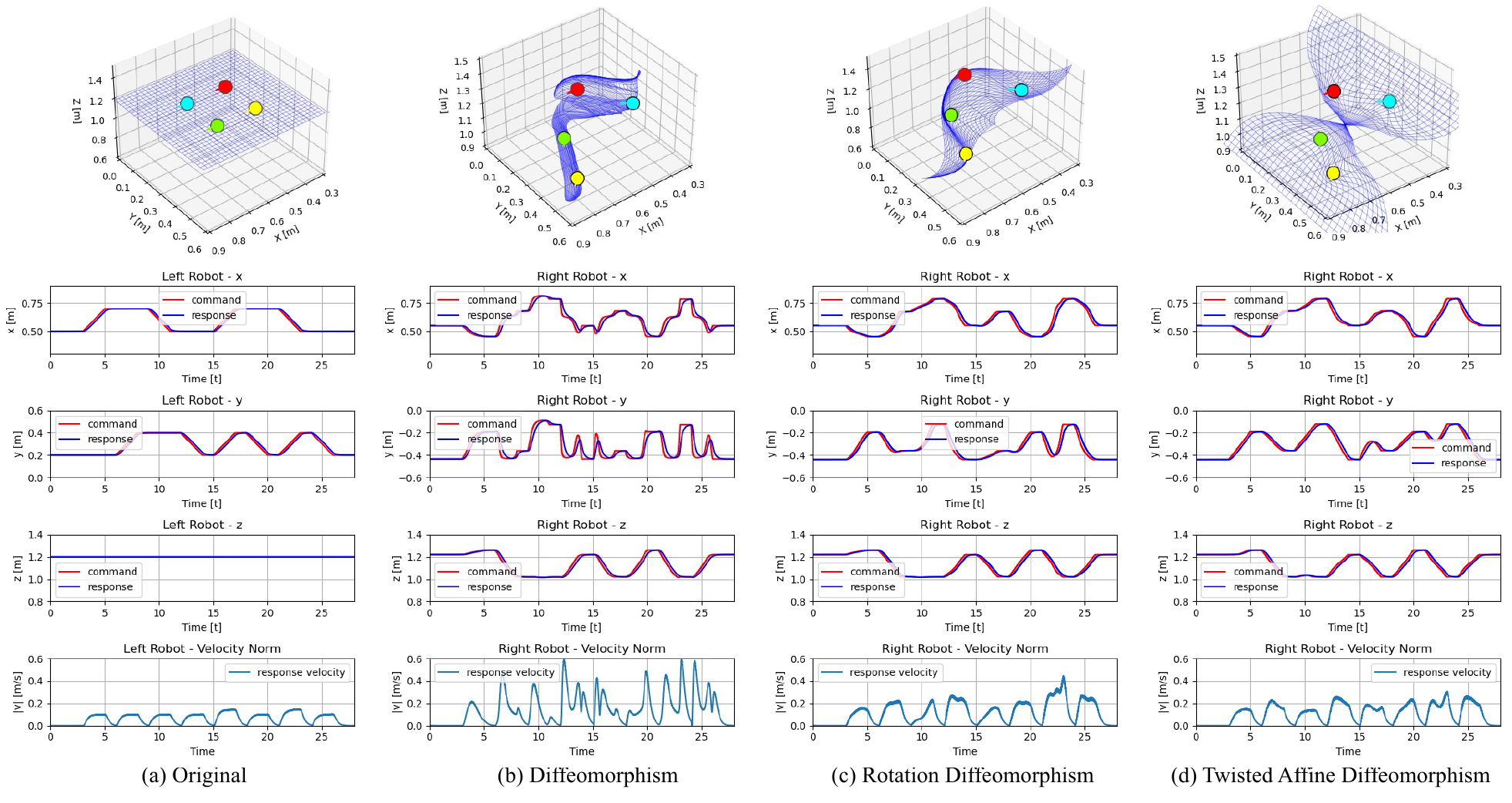}
  \caption{Results of each mapping method and position/velocity response of the robot in Experiment 1.}
  \label{fig:exp1-result}
\end{center}
\end{figure*}

\section{Experiment}
\label{experiment}

\subsection{Setup}
To verify the effectiveness of the proposed method, we conduct experiments using a dual-arm UR5 (Universal Robots) with 2F-85 grippers (Robotiq). A ROS2 velocity controller is implemented for the robot. The experimental setup of the dual-arm UR5 robot is shown in Fig.~\ref{fig:experimental_setup}. The robot is connected to a PC running Ubuntu 22.04 with a PREEMPT\_RT patched kernel. The right arm is offset in the $y$-direction by +0.6\,m relative to the left arm.

Table~\ref{objects_exp} lists the target positions used in each experiment. In Experiment 2, red, yellow, green, and blue blocks are used as target objects representing the robot's environment. Their positions and orientations are detected using AprilTag (AR marker) and a ZED 2i stereo camera. We employ a Leap Motion Controller 2.

\begin{table}[thbp]
\caption{Target positions in each environment}
\begin{center}
\begin{tabular}{ccc}
\hline\hline
  & Right Arm & Left Arm \\
\hline\hline
Experiment 1  & Preset Values & Preset Values \\
\hline
Target 1 ${\color{red}\bm{\bullet}}$ & ${(0.50, 0.20, 1.20)}$ & ${(0.79, 0.48, 1.02)}$\\
Target 2 ${\color{blue}\bm{\bullet}}$ & ${(0.70, 0.20, 1.20)}$ & ${(0.55, 0.16, 1.22)}$\\
Target 3 ${\color{green}\bm{\bullet}}$ & ${(0.70, 0.40, 1.20)}$ & ${(0.45, 0.41, 1.26)}$\\
Target 4 ${\color{yellow}\bm{\bullet}}$ & ${(0.50, 0.40, 1.20)}$ & ${(0.68, 0.24, 1.02)}$\\
\hline\hline
Experiment 2  & Detected (AprilTag)& Detected (AprilTag)\\
\hline
Target 1 ${\color{red}\bm{\bullet}}$ & ${(0.492, 0.502, 0.898)}$ & ${(0.746, 0.344, 1.061)}$\\
Target 2 ${\color{blue}\bm{\bullet}}$ & ${(0.488, 0.163, 0.919)}$ & ${(0.482, 0.334, 0.912)}$\\
Target 3 ${\color{green}\bm{\bullet}}$ & ${(0.721, 0.164, 0.927)}$ & ${(0.849, 0.146, 0.961)}$\\
Target 4 ${\color{yellow}\bm{\bullet}}$ & ${(0.756, 0.506, 0.905)}$ & ${(0.564, 0.059, 0.896)}$\\
\hline\hline
\end{tabular}
\label{objects_exp}
\end{center}
\end{table}

\subsection{Experiment 1}
In Experiment 1, the primary (left) robot follows a trajectory in the order: Target 1 $\to$ 2 $\to$ 3 $\to$ 4 $\to$ 1 $\to$ 3 $\to$ 2 $\to$ 4. We assess the responses of the follower (right) robot under the following mapping methods: DIFF, R-DIFF, and TA-DIFF.

Fig. \ref{fig:exp1-result} shows the command and response position of the end-effector, and the velocity norm for each mapping. In DIFF, oscillatory behavior was observed in the position response when moving from Target 4 to 1. Besides, the generated motion in DIFF environment included a trajectory with a larger velocity norm than in the original environment. In contrast, with the proposed R-DIFF and TA-DIFF, the velocity deviation from the original trajectory was small, and the position response remained smooth. The evaluation indexes for each mapping are shown in Table \ref{tab:exp1_mapping}. The proposed mappings significantly reduced the velocity gradient while achieving the required positional accuracy.
\begin{table}[t]
\caption{Evaluation metrics in Experiment 1}
\begin{center}
\begin{tabular}{cccc}
\hline\hline
 & (a) DIFF& (b) R-DIFF& (c) TA-DIFF\\
\hline
Positional Error& 3.26\,mm& 3.36\,mm&  2.94\,mm\\
Maximum Gradient& 44.13& 5.41& 5.39\\
RBF Iteration& 88& 28& 4 \\
\hline\hline
\vspace{-10mm}
\end{tabular}
\label{tab:exp1_mapping}
\end{center}
\end{table}

\subsection{Experiment 2}
In Experiment 2, the operator conducts a task grasping Targets 1–4 and moving them to the location of Target 2 in both the primary and follower environment at the same time, as shown in Fig. \ref{fig:exp2-result}. The pinch and grab gestures of the operator's hand are used as triggers to activate the robot operation and open/close the mounted gripper, respectively. The operator conducts the task three times.

In the case of DIFF, the follower robot sometimes moved very quickly even though the commanded velocity for the primary robot was not so large. This caused an emergency stop because the follower robot reached its safety limiter.
In the case of R-DIFF, no emergency stop of the follower robot occurred. However, the prediction of the movement of the follower robot would be difficult because the shape of the map is not intuitive to the human operator, which prolongs the task completion.
In the case of TA-DIFF, thanks to the smooth and intuitive shape of the mapping, the operator was able to grasp all targets. 

\section{Discussion}
The proposed R-DIFF is compatible with an infinite number of key points theoretically, which is advantageous when adding multiple sub-key points around target objects, as in simulation 1. Compared to DIFF, R-DIFF tends to improve the mapping gradient, but in some cases, the mapping error is higher than DIFF.
TA-DIFF performs the best in terms of mapping error and mapping gradient, but since it assumes that the number of key points is four, further consideration is needed for more than 5 key points.

\section{Conclusion}

In this paper, we presented the Rotation Extended Diffeomorphism and the Twisted Affine Extended Diffeomorphism, and their application as a mapping method for real-time motion replication. Through simulation, our approach is more compatible than the conventional method in terms of mapping gradient and mapping error. 
The experimental results show that the proposed mappings enabled smooth position mappings for robots in the replicated environment, while maintaining the smallest velocity deviation from a primary robot. 
These findings indicate that the proposed method would be effective for real-time motion reproduction in multi-robot teleoperation.
Future work will address mappings at levels of velocity and acceleration to facilitate smooth environmental interaction of each robot.

\section*{APPENDIX}
\footnotesize
\begin{equation}
   R_{\text{scaling}}(a_x, a_y, a_z) = \mathrm{diag} (a_{x}, a_{y}, a_{z}) \label{equ:scaling}  
\end{equation}
\begin{equation}
R_{\text{shear}}(s_{xy}, s_{xz}, s_{yz}) = \begin{bmatrix}
  1 & s_{xy} & s_{xz} \\
  0 & 1 & s_{yz} \\
  0 & 0 & 1 \\
  \end{bmatrix}
  \label{equ:shear} 
  \end{equation}
  \begin{equation}
  R_{\text{reflection}}(r_x, r_y, r_z) = \rm{diag}(r_x, r_y, r_z) \label{equ:reflection}
    \end{equation}
  \begin{equation}
  R_{\text{rotation}}(\phi, \theta, \psi) = R_{\text{yaw}}(\psi) \cdot R_{\text{pitch}}(\theta) \cdot R_{\text{roll}}(\phi) \label{equ:rotation} 
  \end{equation}
  \begin{equation}
  R_{\text{twist}}(\phi_t(x), \theta_t(y)) = P \cdot R_{\text{pitch}}(\theta_t(y)) \cdot P \cdot  R_{\text{roll}}(\phi_t(x)) + (I - P) \label{equ:twist_rotation} 
    \end{equation}
  \begin{equation}
  P = \mathrm{diag}(1, 1, 0)
 \end{equation}
\normalsize


\end{document}